\documentclass[letterpaper]{article}
\usepackage{proceed2e}
\usepackage[margin=1in]{geometry}

\usepackage{natbib}
\setcitestyle{authoryear,open={(},close={)}}

\usepackage{times}
\usepackage{graphicx}
\usepackage[caption=true]{subfig}
\usepackage[shortlabels]{enumitem}
\usepackage{parskip}
\usepackage{amsmath}
\usepackage{float}
\usepackage{booktabs}
\usepackage{multirow}
\usepackage[table,xcdraw]{xcolor}
\usepackage[hidelinks]{hyperref}
\usepackage{amsthm}
\usepackage{caption}

\usepackage[framemethod=TikZ]{mdframed}

\theoremstyle{definition}
\newtheorem{example}{Example}

\title{Stochastic Layer-Wise Precision in Deep Neural Networks}

\author{ {\bf Griffin Lacey\thanks{*Work completed while at the University of Guelph}} \\
NVIDIA \\
\And
{\bf Graham W. Taylor}  \\
University of Guelph          \\
Vector Institute for Artificial Intelligence\\
Canadian Institute for Advanced Research
\And
{\bf Shawki Areibi}   \\
University of Guelph \\
}

\begin{document}
\maketitle

\begin{abstract}
Low precision weights, activations, and gradients have been proposed as a way to
improve the computational efficiency and memory footprint of deep neural
networks. Recently, low precision networks have even shown to be more robust to
adversarial attacks. However, typical implementations of low precision DNNs use
uniform precision across all layers of the network. In this work, we explore
whether a heterogeneous allocation of precision across a network leads to
improved performance, and introduce a learning scheme where a DNN stochastically
explores multiple precision configurations through learning. This permits a
network to learn an optimal precision configuration. We show on convolutional neural networks
trained on MNIST and ILSVRC12 that even though these nets learn a uniform or
near-uniform allocation strategy respectively, stochastic precision leads to a
favourable regularization effect improving generalization.
\end{abstract}

\section{INTRODUCTION}

Recent advances in deep learning, and convolutional neural networks (CNN) in
particular, have led to well-publicized breakthroughs in computer vision
\citep{Krizhevsky2012-ij}, speech recognition \citep{hinton2012deep}, and
natural language processing (NLP) \citep{bahdanau2014neural}. Modern
CNNs, however, have increasingly large storage and computational requirements
 \citep{DBLP:journals/corr/CanzianiPC16}. This has limited the application scope to data centres
 that can accommodate clusters of massively parallel hardware accelerators, such as
graphics processing units (GPUs). Still, GPU training of CNNs on large datasets
like ImageNet \citep{deng2009imagenet} can take hours, even on networks with
hundreds of GPUs, and achieving linear scaling beyond these sizes is difficult
\citep{goyal2017accurate}. As such, there is a growing interest in
investigating more fine-grained optimizations, especially since deployment on
embedded devices with limited power, compute, and memory budgets remains an
imposing challenge.

Research efforts to reduce model size and speed up inference have shown
that training networks with binary or ternary weights and activations
\citep{DBLP:journals/corr/CourbariauxB16,Rastegari2016-vd,li2016ternary} can achieve
comparable accuracy to full precision networks, while benefiting from reduced
memory requirements and improved computational efficiency using bit
operations. They may even confer additional robustness to adversarial attacks
\citep{Galloway2018-cy}. More recently, the DoReFa-Net model has generalized this finding
to include different precision settings for weights vs.~activations, and demonstrated how
low precision \emph{gradients} can be also employed at training time~\citep{Zhou2016-fh}.

These findings suggest that precision in deep learning is not an arbitrary
design choice, but rather a dial that controls the trade-off between model
complexity and accuracy. However, precision is typically considered at the
design level of an entire model, making it difficult to consider as a
tunable hyperparameter. We posit that considering
precision at a finer granularity, such as a layer or even per-example
could grant models more flexibility in which to find optimal configurations,
which maximizes accuracy and minimizes computational cost. To remain deterministic
about hardware efficiency, we aim to do this for fixed budgets of precision, which have
predictable acceleration properties.


In this work we consider learning an optimal precision configuration across the layers of
a deep neural network, where the precision assigned to each layer may be
different. We propose a stochastic regularization technique akin to
Dropout~\citep{Srivastava2014-dq} where a network explores a different
precision configuration per example. This introduces non-differentiable elements
in the computational graph which we circumvent using recently proposed gradient
estimation techniques.

\section{RELATED WORK}

Recent work related to efficient learning has explored a number of different approaches to reducing
the effective parameter count or memory footprint of CNN architectures. Network compression techniques
\citep{han2015deep, fg2016scalable, choi2016towards, agustsson2017soft} typically compress
a pre-trained network while minimizing the degradation of network accuracy. However, these
methods are decoupled from learning, and are only suitable for efficient deployment. Network
pruning techniques \citep{wan2013regularization, han2015learning, jin2016training,
li2016pruning,
anwar2017structured, molchanov2016pruning,Tung2017-dj} take a more iterative approach, often using
regularized training and retraining to rank and modify network parameters based on their
magnitude. Though coupled with the learning process, iterative pruning techniques tend to
contribute to slower learning, and result in sparse connections, which are not hardware efficient.

Our work relates primarily to low precision techniques, which have tended to focus on reducing the
precision of weights and activations used for deployment while maintaining dense connectivity.
Courbariaux et al.~were among the first to explore binary weights and activations
\citep{DBLP:journals/corr/CourbariauxBD15,DBLP:journals/corr/CourbariauxB16}, demonstrating
state-of-the-art results for smaller datasets (MNIST, CIFAR-10, and CVHN). This idea was then
extended further with CNNs on larger datasets like ImageNet with binary weights and
activations, while approximating convolutions
using binary operations \citep{Rastegari2016-vd}. Related work \citep{kim2016bitwise} has
validated these results and shown neural networks to be remarkably robust to an even wider class
of non-linear projections \citep{merolla2016deep}. Ternary quantization strategies
\citep{li2016ternary} have been shown to outperform their binary counterparts, moreso when
parameters of the quantization module are learned through backpropagation
\citep{Zhu2016-aq}. Cai et al. have investigated how to improve the gradient quality of
quantization operations \citep{halfwave}, which is complimentary to our work which relies on
these gradients to learn precision.  Zhou et al.~further explored this idea of variable precision
(i.e.~heterogeneity across weights and activations) and discussed the general trade-off of 
precision and accuracy, exploring strategies for training with low precision gradients
\citep{Zhou2016-fh}.

Our approach for learning precision closely resembles BitNet \citep{Raghavan2017-ae}, where the
optimal precision for each network layer is learned through gradient descent, and network
parameter encodings act as regularizers. While BitNet uses the Lagrangian approach of adding
constraints on quantization error and precision to
the objective function, we allocate bits to layers through sampling from a Gumbel-Softmax
distribution constructed over the network layers. This has the advantage of accommodating a
defined precision budget, which allows more deterministic hardware constraints, as well as a
wider range of quantization encodings through the use of non-integer quantization values early in
training. In the allocation of bits on a budget, our work resembles \citep{fg2016scalable},
though we allow more fine-grained control over precision, and prefer a gradient-based approach
over clustering techniques for learning optimal precision configurations.

To the best of our knowledge, our work is the first to explore learning precision in deep networks through a continuous-to-discrete annealed quantization strategy. Our contributions are as follows:

\begin{itemize}
\item We experimentally confirm a linear relationship between total number of bits and speedup
for low precision arithmetic, motivating the use of precision budgets.
\item We introduce a gradient-based approach to learning precision through sampling from a
Gumbel-Softmax distribution constructed over the network layers, constrained by a precision
budget.
\item We empirically demonstrate the advantage of our end-to-end training strategy as it
improves model performance over simple uniform bit allocations.
\end{itemize}

\section{EFFICIENT LOW PRECISION NETWORKS}

Low precision learning describes a set of techniques that take network parameters, typically
stored at native 32-bit floating point (FP32) precision, and quantize them to a much smaller
range of representation, typically 1-bit (binary) or 2-bit (ternary) integer values. While low precision
learning could refer to any combination of quantizing weights (W), activations (A), and gradients
(G), most relevant work investigates the effects of quantizing weights and activations on model
performance. The benefits of quantization are seen in both computational and memory efficiency,
though generally speaking, quantization leads to a decrease in model accuracy
\citep{Zhou2016-fh}. However, in some cases, the effects of quantization can be lossless or
even slightly improve accuracy by behaving as a type of noisy regularization
\citep{Zhu2016-aq,Yin2016-pk}. In this work, we adopt the DoReFa-Net model \citep{Zhou2016-fh}
of quantizing all
network parameters (W,A,G) albeit at different precision. Table \ref{tab:dorefa-accuracy} demonstrates this
trade-off of precision and accuracy for some common low precision configurations.

\begin{table*}
\rowcolors{1}{white}{lightgray}
\centering
\caption{DoReFa-Net single-crop top-1 validation error for common weight (W), activation (A), and
gradient (G)
quantization configurations on the ImageNet Large-Scale Visual Recognition Challenge 2012 (ILSVRC12) dataset.  The results are slightly
improved over the results originally reported and were reproduced by us based on the public DoReFa-Net \citep{Zhou2016-fh}
codebase.}
\label{tab:dorefa-accuracy}
\begin{tabular}{lcccc}
\toprule
Model			&W	& A		& G		& Top-1 Validation Error 	\\ \midrule
AlexNet	\citep{Krizhevsky2012-ij}		&32	& 32  	& 32		& 41.4\%		\\
BWN	\citep{DBLP:journals/corr/CourbariauxB16}			&1 	& 32  	& 32		& 44.3\%
\\
DoReFa-Net	\citep{Zhou2016-fh}	&1 	& 2		& 6		& 47.6\% 		\\
DoReFa-Net	\citep{Zhou2016-fh}	  &1 	& 2		& 4		& 58.4\% 		\\ \bottomrule
\end{tabular}
\end{table*}

The justification for this loss in accuracy is the efficiency gain of storing and computing low
precision values. The computational benefits of using binary values are seen from
approximating expensive full precision matrix operations with binary operations
\citep{Rastegari2016-vd}, as well as reducing memory requirements by packing many low
precision values into full precision data types. For other low precision configurations that fall
between binary and full precision, a similar formulation is used. The bit dot product equation
(Equation \ref{eq:bitdot}) shows how both the logical \texttt{and} and \texttt{bitcount} operations
are used to compute the dot product of two low-bitwidth fixed-point integers (bit vectors)
\citep{Zhou2016-fh}. Assume $c_m(x)$ is a placewise bit vector formed from a sequence of
$M$-bit fixed-point integers $x=\sum_{m=0}^{M-1}c_m(x)2^m$ and $c_k(y)$ is a placewise bit
vector formed from a sequence of $K$-bit fixed-point integers $y=\sum_{k=0}^{M-1}c_k(y)2^k$,
then
\begin{align}
\label{eq:bitdot}
\!\!\!x \cdot y \! & = \!\!
\sum^{M-1}_{m=0} \sum^{K-1}_{k=0} 2^{m+k} \,
\text{bitcount} [\text{and}(c_{m}(x),c_{k}(y))],\nonumber \\
& c_{m}(x)_i,c_{k}(y)_i \in \{0,1\} \forall i,m,k.
\end{align}

\begin{figure}
	\centering
	\includegraphics[width=3.20in]{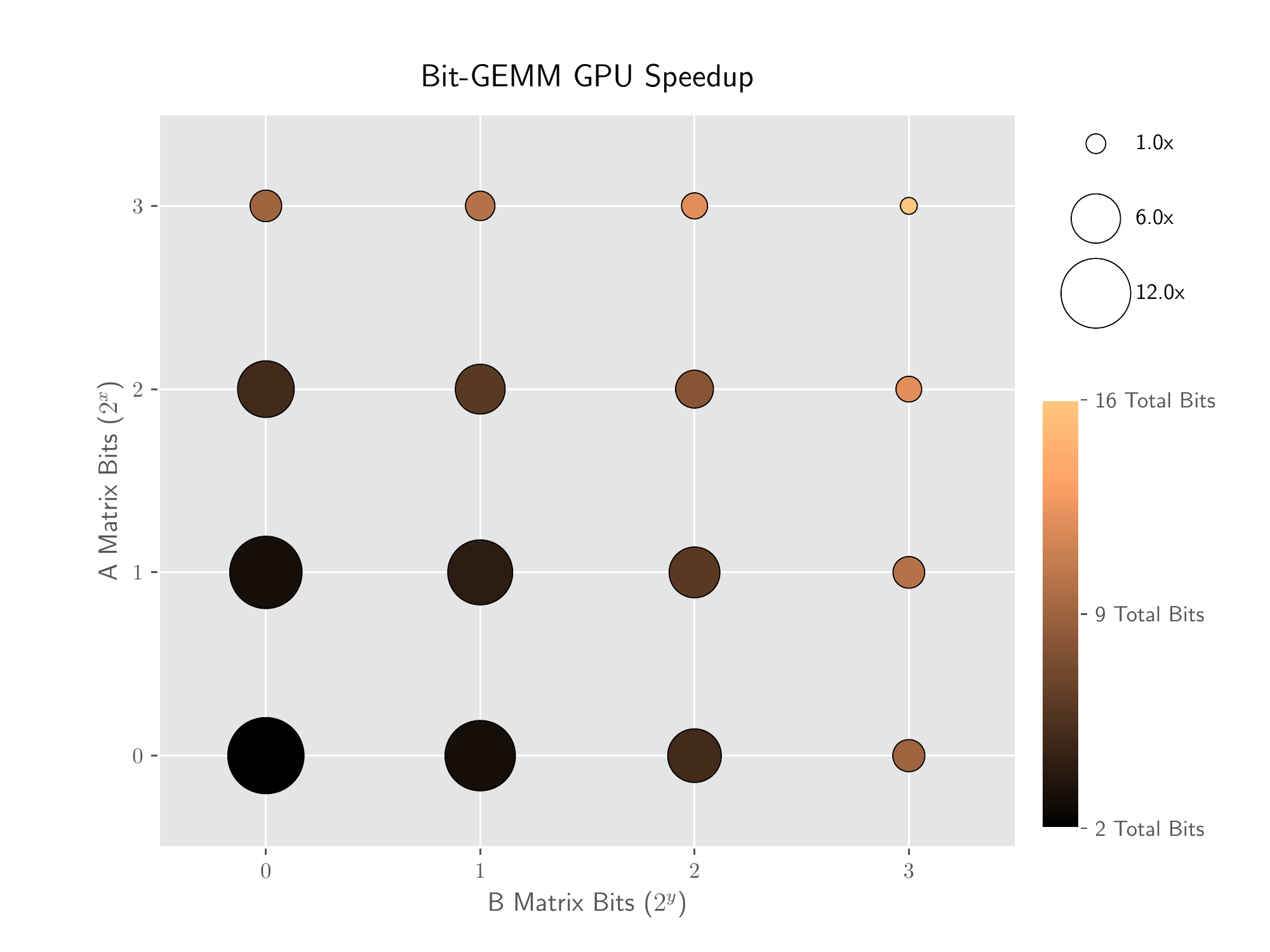}
	\caption{GPU-based bit-GEMM speedup for low precision matrices A and B. Results are
	compared with a similarly optimized 32-bit GEMM kernel, and run on a NVIDIA Tesla V100
	GPU.}
	\label{fig:GPUbitgemm}
\end{figure}

Since the computational complexity of the operation is $\mathcal{O}(MK)$, the speedup is a function of the
total number of bits used to quantize the inputs. Since matrix multiplications are simply
sequences of dot products computed over the rows and columns of matrices, this is also true of
matrix multiplication operations. As a demonstration, we implement
this variable precision bit general matrix multiplication (bit-GEMM) using CUDA, and show the
GPU speedup for several configurations in Figure \ref{fig:GPUbitgemm}.

As seen in Figure \ref{fig:GPUbitgemm}, there is a correlation between the total number
of bits used in each bit-GEMM (shade) and the resulting speedup (point size). As such, from a
hardware perspective, it is important to know how many total bits are used for bit-GEMM
operations to allow for budgeting computation and memory. It should also be noted that, in our
experiments, operations with over 16 total bits of precision were shown to be slower than the full
precision equivalent.  This is due to the computational complexity of the worst case of $\mathcal{O}(8\times8)$
being slower than the equivalent full precision operation.  We therefore focus on
operations with 16 or less total bits of precision.

A natural question to follow is then, for a given budget of precision (total
number of bits), how do we most efficiently allocate precision to maximize model
performance? We seek to answer this question by parametrizing the precision
at each layer and learning these additional parameters by gradient descent.

\subsection{LEARNING PRECISION}

\begin{figure}
 	\centering
 	\includegraphics[width=3.20in]{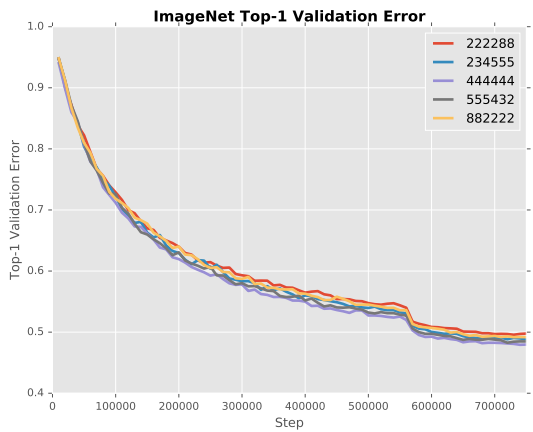}
 	\caption{The training error for a variety of DoReFa-Net manual precision allocations is plotted
  for each weight update.
 	The uniform distribution of bits (i.e.~\textbf{444444}) leads to the lowest error, while the
  least uniform configurations (e.g.~\textbf{222288}, \textbf{882222}) lead to the highest error.}
 	\label{manual}
 \end{figure}

The selection of precision for variables in a model can have a significant impact on performance.
Consider the DoReFa-Net model --- if we decide on a budget of the total number of bits to assign
to the weights layer-wise, and train the model under a number of different manual allocations,
we obtain the training curves in Figure \ref{manual}. For each training curve, the number of bits
assigned to each layer's weights are indicated by the integer at the appropriate position (e.g.\
\textbf{444444} indicates 6 quantized layers, all assigned 4 bits).

 Varying the number of bits assigned to each layer can cause the error to change by up to
 several percent, but the best configuration of these bits is unclear without exhaustively
 testing all possible configurations. This motivates \emph{learning} the most efficient allocation
 of precision to each layer. However, this is a difficult task for two important reasons:

\begin{itemize}
\item unconstrained parameters that control precision will only grow, as higher precision leads
  to a reduction of loss; and
\item quantization involves discrete operations, which are non-differentiable and therefore
unsuitable for na\"ive backpropagation.
\end{itemize}

The first issue is easily addressed by fixing the total network precision
(i.e.~the sum of the bits of precision at each layer) to a budget $B$, similar
to the $B=24$ configurations seen in Figure~\ref{manual}. The task is then to
learn the allocation of precision across layers.
The second issue: the non-differentiable nature of quantization operations
is an unavoidable problem, as transforming continuous values into discrete
values must apply some kind of discrete operator. We avoid this issue by
employing a kind of stochastic allocation of
precision and rely on recently developed techniques from the deep learning
community to backpropagate gradients through discrete stochastic operations.

Intuitively, we can view the precision allocation procedure as sequentially
allocating one bit of precision to one of $L$ layers. Each time we allocate, we
draw from a categorical variable with $L$ outcomes, and allocate that bit to the
corresponding layer. This is repeated $B$ times to match the precision budget.
An allocation of $B$ bits corresponds to a particular precision configuration,
and we sample a new configuration for each input example.  The idea of
stochastically sampling architectural configurations is akin to Dropout
\citep{Srivastava2014-dq}, where each example is processed by a different
architecture with tied parameters.

Different from Dropout, which uses a fixed dropout probability, we would like
to parametrize the categorical distribution across layers such that we can
learn to prefer to allocate precision to certain layers. Learning these
parameters by gradient descent requires backprop through an operator that
samples from a discrete distribution. To deal with its non-differentiability,
we use the Gumbel-Softmax, also known as the Concrete distribution \citep{Jang2016-qi,
Maddison2016-ab}, which, using a temperature parameter, can be smoothly annealed from a uniform
continuous distribution on the simplex to a discrete categorical distribution from which we sample precision allocations. This
allows us to use a high temperature at the beginning of training to stochastically explore
different precision configurations and use a low temperature at the end of training to discretely
allocate precision to network layers according to the learned distribution.

Though non-integer bits of precision can be implemented (detailed below),
integer bits are more amenable to hardware implementations, which is why we aim to converge toward
discrete samples.
Table \ref{tab:gumbelsoftmax} shows
examples of sampling from this distribution at different temperatures for a three class
distribution. It should be noted that in order to perform unconstrained optimization we
parametrize the unnormalized logits (also known as log-odds) instead of the probabilities themselves.

\begin{table}[h!]
	\rowcolors{1}{white}{lightgray}
	\centering
	\caption{Examples of single samples drawn from a Gumbel-Softmax distribution, parametrized
	by logits for three classes $\pi_1, \pi_2, \pi_3$ and a variety of temperatures $\tau$. The
	probability of allocating a bit to layer $l_i$ is impacted by the logit $\pi_i$, where higher values
	correspond to higher probabilities. The Gumbel-Softmax interpolates between continuous densities
  on the simplex (at high temperature) and discrete one-hot-encoded categorical distributions (at low temperature).}
	\label{tab:gumbelsoftmax}
	\begin{tabular}{cccc}
		\toprule
		$\tau$	& \shortstack{Class 1 \\ ($\pi_1 = 1.00$)} & \shortstack{Class 2 \\
		($\pi_2 = 2.00$)} 	& \shortstack{Class 3 \\ ($\pi_3 =	-0.50$)}	 \\\midrule
		100.0     & 0.33  	& 0.33	  & 0.33  		\\
		10.0     	& 0.33  	& 0.41	  & 0.26  		\\
		1.00     	& 0.31  	& 0.60	  & 0.09  		\\
		0.10     	& 0.00  	 & 1.00	  & 0.00  		\\\bottomrule
	\end{tabular}
\end{table}

Since the class logits $\pi_i$ control the probability of allocating a bit of precision to a
network layer $l_i$, at low temperatures the one-hot samples will allocate bits of precision to the
network according to the learned parameters $\pi_i$. However, at high temperatures we
allocate partial bits to layers. This is possible due to our quantization straight-through
estimator (STE), \texttt{quantize}, adopted from \citep{Zhou2016-fh}:
\begin{align}
\label{eq:STE}
&\textbf{Forward: } \nonumber \\
&r_o\!=\!\texttt{quantize}(r_i)\!=\!\frac{1}{2^k-1}\text{round}\left((2^k-1)r_i\right),\\
&\textbf{Backward: } \frac{\partial c}{\partial r_i} = \frac{\partial c}{\partial r_o}
\end{align}
where $r_i$ is the real number input, $r_o$ is the $k$-bit output, and $c$ is the objective
function. Since \texttt{quantize} produces a $k$-bit value on $[0,1]$, quantizing to non-integer
values of $k$ simply produces a more fine-grained range of representation compared to integer
values of $k$. This is demonstrated in Table \ref{tab:nonintegerk}.

\begin{table}[h!]
	\rowcolors{1}{white}{lightgray}
	\centering
	\caption{The possible output values of the \texttt{quantize} operation are shown for a variety of
	values 	of $k$, clipped between 0 and 1. Non-integer values of $k$ provide a more
	fine-grained range of representation between successive integer values (underlined).}
	\label{tab:nonintegerk}
	\begin{tabular}{cccccccc}
		\toprule
		$k$	& \underline{1} & $1.50$ & \underline{2} & $2.25$ & $2.50$ & $2.75$ & \underline{3} 	\\ \midrule
		& 0.00 & 0.00 & 0.00 & 0.00 & 0.00 & 0.00 & 0.00\\
		& 1.00 & 0.55 & 0.33 & 0.26 & 0.21 & 0.17 & 0.14\\
	 & &  1.00 & 0.66 & 0.53 & 0.42 & 0.34 & 0.28\\
		& & &    1.00 & 0.80 & 0.64 & 0.52 & 0.42\\
		& & & &    1.00 & 0.85 & 0.69 & 0.57\\
		& & & & &     1.00 & 0.87 & 0.71\\
		& & & & & &       1.00 & 0.85\\
		& & & & & & &        1.00\\ \bottomrule
	\end{tabular}
\end{table}

Using real values for quantization also provides useful gradients for backpropagation, whereas
the small finite set of possible integer values would yield zero gradient almost everywhere.


\subsection{PRECISION ALLOCATION LAYER}

\begin{figure*}[]
 	\centering
 	\includegraphics[width=\textwidth]{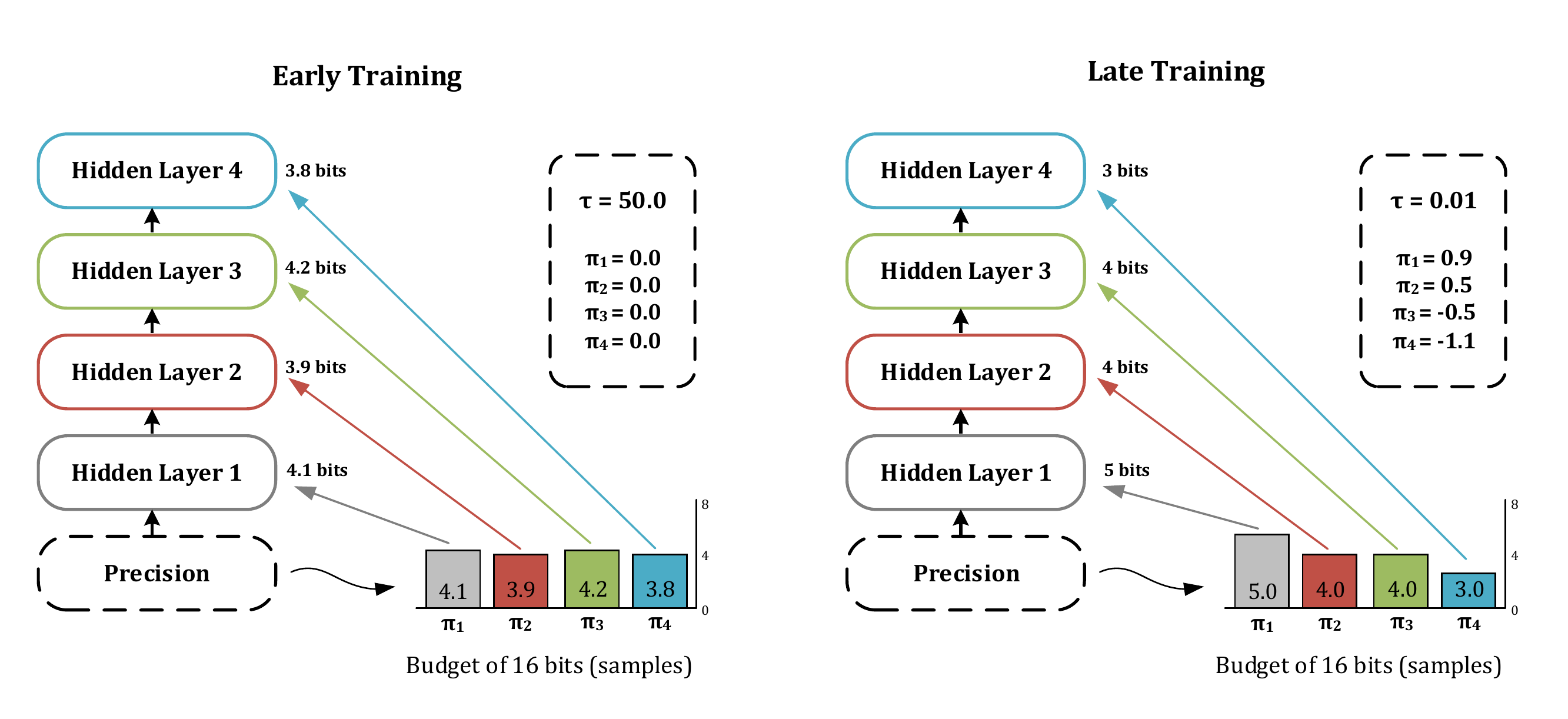}
 	\caption{Early in training, the Gumbel-Softmax class probabilities $\pi_i$ are initialized
 	to 0 while the temperature $\tau$ is high, generally resulting in uniform allocations of
 	real-valued bits of precision. Later in training, with a low temperature, the learned class
 	probabilities usually result in some layers being allocated more or less discrete bits of
 	precision.  In practice, the computational overhead of the precision layer is not noticeable.}
 	\label{fig:prec}
 \end{figure*}

We introduce a new layer type, the precision allocation layer, to implement our precision learning
technique. This layer is inserted as a leaf in the computational graph, and is executed on each
forward-pass of training. The precision allocation layer is parametrized by the learnable class
probabilities $\pi_i$, which define the Gumbel-Softmax distribution. Each class probability is
associated with a layer $l_i$, so the samples assigned to each class are allocated as bits to
the appropriate layer. This is illustrated in Figure \ref{fig:prec} and stepped through in
Example~\ref{exa:training}.

\begin{figure*}
	\begin{mdframed}[roundcorner=10pt,backgroundcolor=white]
	\begin{example}\label{exa:training}
	\small
	Consider a network with 4 layers, each denoted by $l_i$.  To learn the precision of these
	layers,
	we add a precision layer which constructs a Gumbel-Softmax distribution with 4 classes
	$\pi_i$, where each class is assigned to a layer.  Each class probability is initalized to 0.0, and
	the temperature is initialized to 50.0.	For a budget of 16 total bits, two example iterations
	representative of early training (first iteration of epoch 0) and late training (first iteration of
	epoch 50) are shown below:

	\textbf{Epoch = 0}, $\pi_1 = 0.0$ , $\pi_2 = 0.0$, $\pi_3 = 0.0$, $\pi_4 = 0.0$, $\tau =
	50.0$
  \begin{itemize}
	\item The precision layer is executed first, which means we sample from the Gumbel-Softmax
	16 times because our budget is 16 bits, and accumulate the results of the 16 samples.  Each
	individual sample gives us 4 class outputs which sum to 1 (e.g.~[0.25, 0.25, 0.25, 0.25]), so
	sampling 16 times and accumulating the results for each class means our final results will add
	to 16.  Since the class probabilities are initialized to
	0.0, and the temperature is high, the expected samples will be continuous and relatively uniform
	across all classes.
    \begin{itemize}
  	\item The samples associated with each layer $l_i$ are: $l_1 = 4.1$ , $l_2 = 3.9$, $l_3 = 4.2$, $l_4 = 3.8$.
  	\item We now assign 4.1 bits to layer 1, 3.9 bits to layer 2, 4.2 bits to layer 3, and 3.8 bits to
  	layer 4.  These bits are assigned to the appropriate layer by applying the \texttt{quantize}
  	operation of Equation \ref{eq:STE} to the desired parameters (e.g.~weights) with $k$ as the
  	appropriate bit assignment (e.g. $k$ = 4.1 for $l_1$).
  	\item The \texttt{quantize} operation will transform the layer parameters to one of several
  	discrete positive quantities between 0 and 1 (see Table \ref{tab:nonintegerk}).  Though
  	these are fractional bit assignments, the \texttt{quantize} operations works the same as if
  	these were integer bit assignments.
  	\item The iteration then proceeds as normal, with the quantized parameters and class
  	probabilities $\pi_i$ updated during back-propagation.
    \end{itemize}
  \end{itemize}

  \textbf{Epoch = 50}, $\pi_1 = 0.9$ , $\pi_2 = 0.5$, $\pi_3 = -0.5$, $\pi_4 = -1.1$, $\tau =
	0.01$
  \begin{itemize}
	\item Similar to before, we begin by sampling from the Gumbel-Softmax 16 times because our
	budget is 16 bits, and accumulate the results.  However, the class probabilities have now
	changed such that $\pi_1$ corresponds to the most likely class, and $\pi_4$ the least likely class,
	so the distribution is no longer uniform.  As well, since the temperature is
	low ($\tau = 0.01$), the samples will now approach discrete.
  \begin{itemize}
  	\item Samples: $l_1 = 5.0$ , $l_2 = 4.0$, $l_3 = 4.0$, $l_4 = 3.0$.
  	\item We now assign 5.0 bits to layer 1, 4.0 bits to layer 2, 4.0 bits to layer 3, and 3.0 bits to
  	layer 4, similar to before.  Since these bit assignments are integer values, the activity of the
  	\texttt{quantize} operation is more intuitive.
  	\item Similar to before, the \texttt{quantize} operation will transform the layer
  	parameters to one of several discrete positive quantities between 0 and 1 (see Table
  	\ref{tab:nonintegerk}).  
  	\item The forward-pass then proceeds as normal, with the quantized parameters and class
  	probabilities $\pi_i$ updated during back-propagation.  Since this is late training, parameter
  	updates will be smaller in magnitude than in early training.
  \end{itemize}
\end{itemize}
\end{example}
\end{mdframed}
\end{figure*}

It should be noted that during the early stages of training before the network performance has
converged, allowing the temperature to drop too low results in high-variance gradients while also
encouraging largely uneven allocations of bits. This severely hurts the generalization of the
network. To deal with this, we empirically observe a temperature where the class probabilities
have sufficiently stabilized, and perform hard assignments of bits of precision based on these
stabilized class probabilities.  To do this, we sample from the Gumbel-Softmax a large number of
times and average the results, in order to converge on the expected class sample assignments.
Once we have these precision values, we fix the layers at these precision values for the remainder
of training.  We observe that the regularization effects of stochastic bit allocations are most
useful during early training, and performing hard assignments greatly improves generalization
performance.  For all experiments considered, we implement a hard assignment of bits after the
temperature drops below $3.0$.

\section{EXPERIMENTS}

We evaluate the effects of our precision layer on two common image
classification benchmarks, MNIST and ILSVRC12. We consider two separate CNN
architectures, a 5-layer network similar to LeNet-5 \citep{Lecun1998-ls} trained
on MNIST, and the AlexNet network \citep{Krizhevsky2012-ij} trained on ILSVRC12.
We use the top-1 single-crop error as a measure of performance, and quantize both
the weights and activations in all experiments considered.  As in previous
works \citep{Zhou2016-fh}, the first layer of AlexNet was not quantized. Initial experiments
showed that the effects of learning precision were less beneficial to gradients,
so we leave them at full precision in all reported experiments.


Since our motivation is to show the precision layer as an improvement over
uniformly quantized low precision models, we compare our results to networks
with evenly distributed precision over all layers. We consider three common low
precision budgets as powers of 2, which would provide efficient hardware
acceleration, where each layer is allocated 2, 4, or 8 bits. Baseline networks
with uniform precision allocation are denoted with the allocation for each layer
and the budget (e.g.~\textbf{budget=10\_22222} denotes a baseline network with 2
bits manually allocated to each of 5 layers for a budget of 10 bits), while the
networks with learned precision are denoted with the precision budget and \textit{learn}
(e.g.~\textbf{budget=10\_learn} denotes a learned precision
network with a budget of 10 total bits, averaging 2 bits per layer).

\subsection{MNIST}

\begin{figure}
 	\centering
 	\includegraphics[width=3.25in]{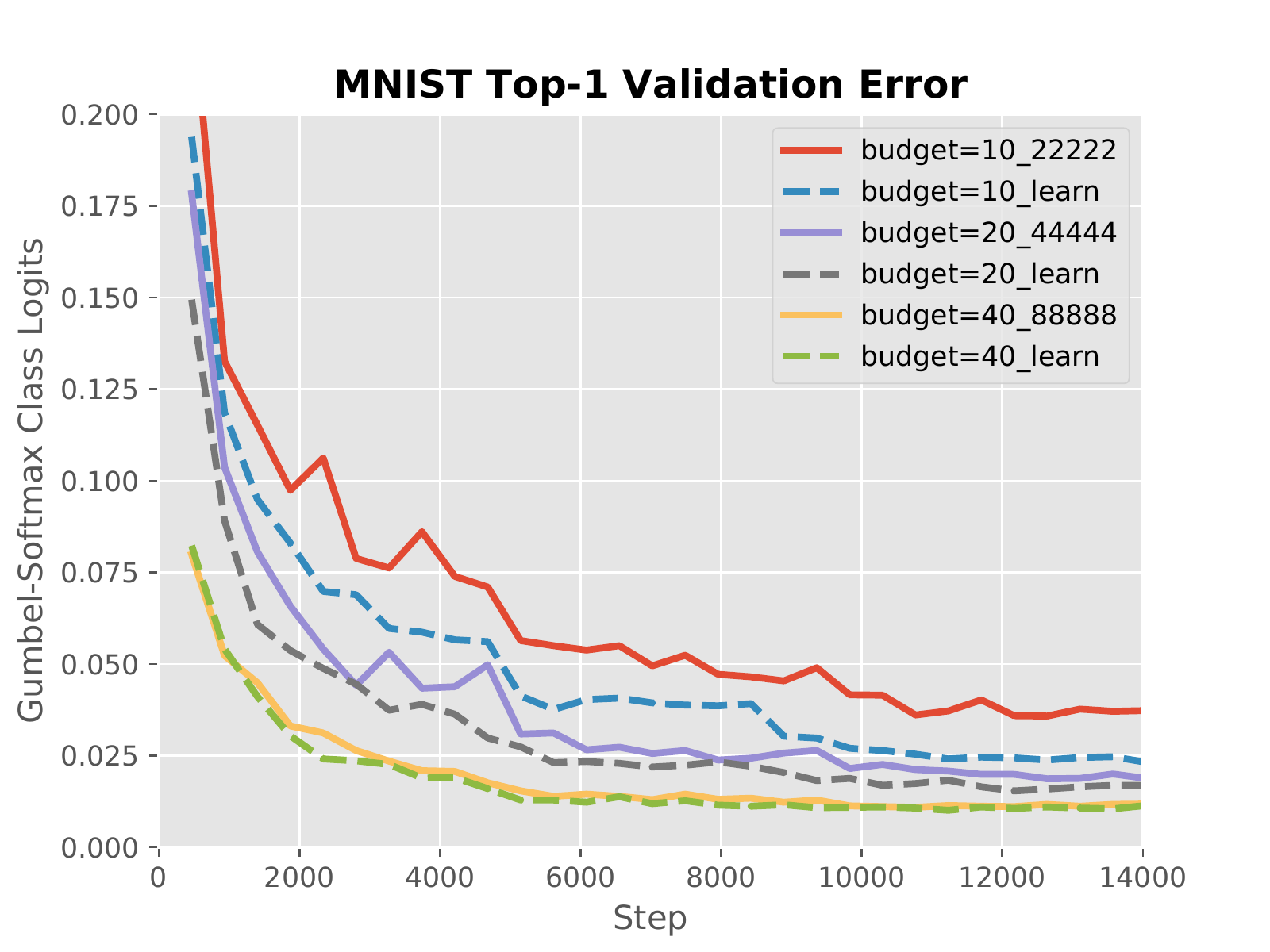}
 	\caption{The top-1 errors for the MNIST networks are shown. Both baseline
 	networks with uniform precision allocation (i.e.~\textbf{budget=10\_22222}) and learned
 	precision networks
 	(i.e.~\textbf{budget=10\_learn}) are assigned the same total precision budget, indicated by the
  prefix.}
 	\label{fig:MNIST_val_acc}
 \end{figure}

\begin{figure*}
 	\centering
 	\includegraphics[width=1\textwidth]{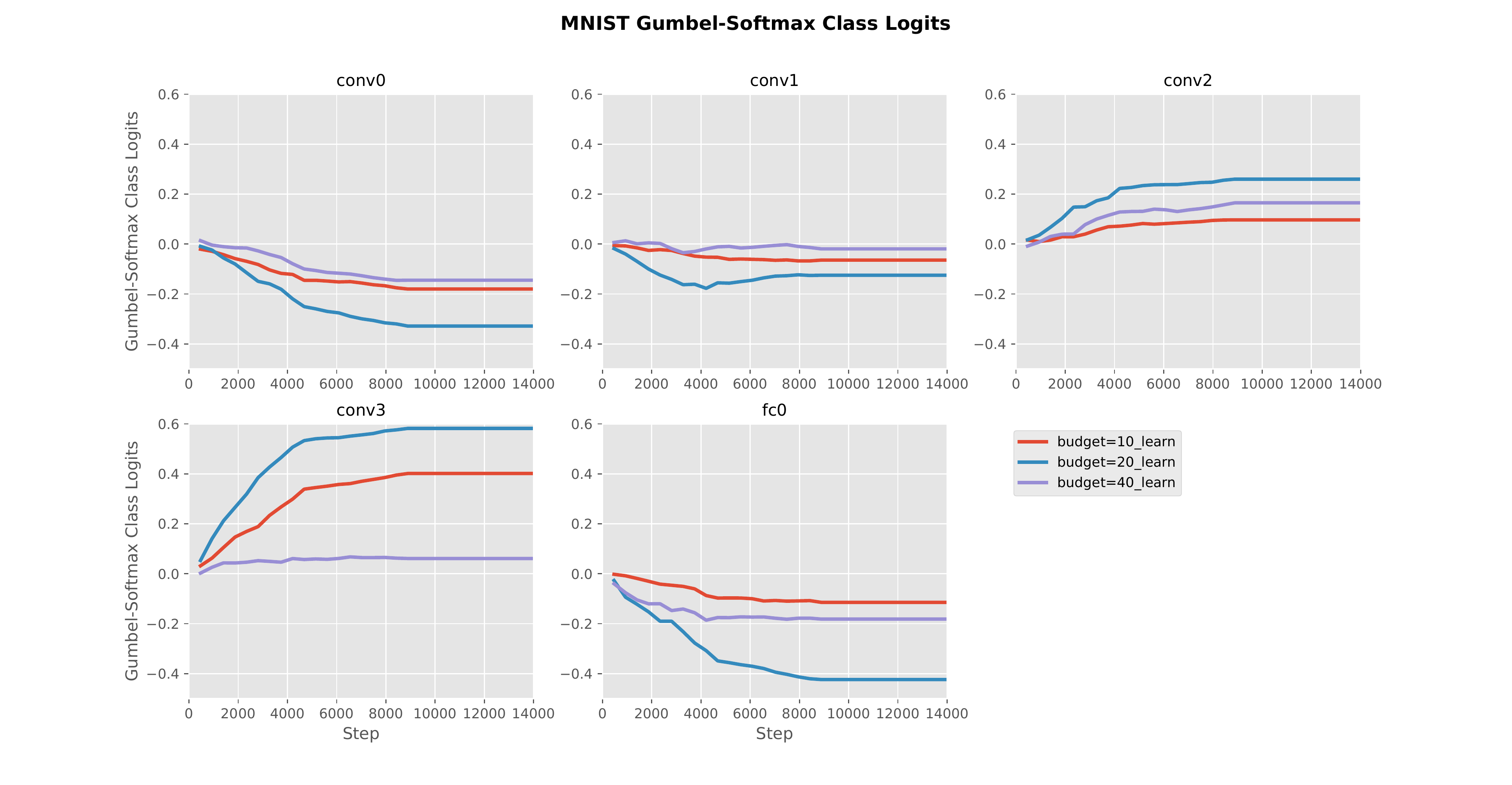}
 	\caption{The learned Gumbel-Softmax class logits for different precision budgets.}
 	\label{fig:gumbel}
\end{figure*}

The training curves for the MNIST-trained models are shown in Figure
\ref{fig:MNIST_val_acc}. The learned Gumbel-Softmax class logits are shown in
Figure \ref{fig:gumbel}.


We observe that the models with learned precision converge faster and reach a lower test error
compared to the baseline models across all
precision budgets considered, and the relative improvement is more substantial for lower
precision budgets. From Figure~\ref{fig:gumbel}, we observe that the models learn to assign fewer
bits to early layers of the network (conv0, conv1) while assigning more bits to later layers of the
network (conv2, conv3), as well as preferring smoother (i.e.~more uniform) allocations. This result
agrees with the empirical observations of \citep{Raghavan2017-ae}.
The results on MNIST are summarized in Table \ref{tab:MNIST}.  Uncertainty is calculated by averaging over 10 runs for each network
with different random initializations of the parameters.


\begin{table*}
	\rowcolors{3}{white}{lightgray}
	\centering
	\caption{Final training results and top-1 error for the MNIST
	baseline and learned precision models.}
	\label{tab:MNIST}
	\begin{tabular}{lrrccccl}
		\toprule
		\multirow{2}{*}{Network} & \multicolumn{1}{r}{\multirow{2}{*}{Precision Budget}} &
		\multirow{2}{*}{Val. Error} & \multicolumn{5}{c}{Final Bit Allocation} \\
		& \multicolumn{1}{l}{} & & \multicolumn{1}{c}{$l_1$} & \multicolumn{1}{c}{$l_2$} &
		\multicolumn{1}{c}{$l_3$} & \multicolumn{1}{c}{$l_4$} & \multicolumn{1}{l}{$l_5$} \\
		\midrule
		budget=10\_22222 & 10 Bits & $3.73 \% \pm 0.3$ & $2$ & $2$ & $2$ & $2$ & $2$ \\
		budget=10\_learn & 10 Bits & $2.32 \% \pm 0.3$ & $2$ & $2$ & $2$ & $2$ & $2$ \\
		budget=20\_44444 & 20 Bits & $1.88 \% \pm 0.2$ & $4$ & $4$ & $4$ & $4$ & $4$ \\
		budget=20\_learn & 20 Bits & $1.69 \% \pm 0.2$ & $4$ & $4$ & $4$ & $4$ & $4$ \\
		budget=40\_88888 & 40 Bits & $1.18 \% \pm 0.1$ & $8$ & $8$ & $8$ & $8$ & $8$ \\
		budget=40\_learn & 40 Bits & $1.14 \% \pm 0.1$ & $8$ & $8$ & $8$ & $8$ & $8$ \\
		\bottomrule
	\end{tabular}
\end{table*}

\begin{table*}
	\rowcolors{3}{white}{lightgray}
	\centering
	\caption{Final training results and top-1error for the ILSVRC12
	baseline and learned precision models.}
	\label{tab:ImageNet}
	\begin{tabular}{lrrccccccl}
		\toprule
		\multirow{2}{*}{Network} & \multicolumn{1}{r}{\multirow{2}{*}{Precision Budget}} &
		\multirow{2}{*}{Val. Error} & \multicolumn{7}{c}{Final Bit Allocation} \\
		& \multicolumn{1}{l}{}  & & \multicolumn{1}{c}{$l_1$} & \multicolumn{1}{c}{$l_2$} &
		\multicolumn{1}{c}{$l_3$} & \multicolumn{1}{c}{$l_4$} & \multicolumn{1}{l}{$l_5$} &
		\multicolumn{1}{l}{$l_6$} & \multicolumn{1}{l}{$l_7$} \\
		\midrule
		budget=14\_2222222 & 14 Bits & $49.68 \% \pm 0.3$ & $2$ & $2$ & $2$ & $2$ & $2$ & $2$
		& $2$ \\
		budget=14\_learn & 14 Bits & $48.54 \% \pm 0.3$ & $2$ & $2$ & $2$ & $2$ & $2$ & $2$ &
		$2$ \\
		budget=28\_4444444 & 28 Bits & $47.99 \% \pm 0.3$ & $4$ & $4$ & $4$ & $4$ & $4$ & $4$
		& $4$ \\
		budget=28\_learn & 28 Bits & $47.69 \% \pm 0.3$ & $3$ & $3$ & $4$ & $4$ & $5$ & $5$ &
		$4$ \\
		budget=56\_8888888 & 56 Bits & $47.70 \% \pm 0.3$ & $8$ & $8$ & $8$ & $8$ & $8$ & $8$
		& $8$ \\
		budget=56\_learn & 56 Bits & $47.46 \% \pm 0.3$ & $7$ & $7$ & $8$ & $8$ & $9$ & $9$ &
		$8$ \\ \bottomrule
	\end{tabular}
\end{table*}

\subsection{ILSVRC12}

The results for ILSVRC12 are summarized in Table~\ref{tab:ImageNet}.
Again, models that employ stochastic precision allocation converge faster and
ultimately reach a lower test error than their fixed-precision counterparts on
the same budget.
We observe that networks trained with stochastic precision
learn to take bits from early
layers and assign these to later layers, similar to the MNIST results. While the
class logits for the MNIST network were similar to the ILSVRC12 results, they
were not substantial enough to cause changes in bit allocation during our hard
assignments. However, the ILSVRC12-trained networks actually make non-uniform
hard assignments. This suggests that the precision layer has a larger affect on
more complex networks.

%
%
%
%

\section{CONCLUSION AND FUTURE WORK}
\label{sec:conclusions}

We introduced a precision allocation layer for DNNs, and proposed a stochastic
allocation scheme for learning precision on a fixed budget.
We have shown that learned precision models outperform uniformly-allocated low
precision models. This effect is due to both learning the
optimal configuration of precision layer-wise, as well as the regularization
effects of stochastically exploring different precision configurations during
training. Moreover, the use of precision budgets allow a high level of
hardware acceleration determinism which has practical implications.

While the present experiments were focused on accuracy rather than computational
efficiency, future work will examine using GPU bit kernels in place of the full
precision kernels we used in our experiments. We also intend to investigate
stochastic precision in the adversarial setting. This is inspired by
\citet{Galloway2018-cy}, who report that stochastic quantization at test time
yields robustness towards iterative attacks.

Finally, we are interested in a variant of the model where rather than directly
parametrizing precision, precision is conditioned on the input. While this
reduces hardware acceleration determinism in real-time or memory-constrained
settings, it would enable a DNN to dynamically adapt its precision configuration
to individual examples.

\bibliographystyle{abbrvnat}
\bibliography{references}

\end{document}